\documentclass[letterpaper, 10 pt, conference]{ieeeconf}  

\IEEEoverridecommandlockouts                              
\usepackage{cite}
\usepackage{amsmath,amssymb,amsfonts}
\usepackage{gensymb}
\usepackage{algorithm}
\usepackage{algpseudocode}
\usepackage{booktabs}
\usepackage{multirow}
\usepackage{graphicx}
\graphicspath{{../pdf/}{../jpeg/}{../Figures/}}
\DeclareGraphicsExtensions{.pdf,.jpeg,.png}

\usepackage{hyperref}
\usepackage{mathtools}

\usepackage{xfrac}

\usepackage[normalem]{ulem}

\newcommand{\IfCondFalse}[1]{%
  \ifnum\pdfstrcmp{\cond}{True}=0 \unskip\else #1\fi\ignorespaces}

\makeatletter
\newcommand{\mypm}{\mathbin{\mathpalette\@mypm\relax}}
\newcommand{\@mypm}[2]{\ooalign{%
  \raisebox{.1\height}{$#1+$}\cr
  \smash{\raisebox{-.6\height}{$#1-$}}\cr}}
\makeatother

\usepackage{textcomp}
\usepackage{xcolor}
\usepackage{subcaption}

\newcounter{todocounter}

\usepackage{hyperref}

\title{\LARGE \bf
Proactive tactile exploration for object-agnostic shape reconstruction from minimal visual priors
}

\author{Paris~Oikonomou$^{1,2,\ast}$,
  ~George~Retsinas$^{1,\ast}$,
  ~Petros~Maragos$^{1,2}$
  ~and~Costas~S.~Tzafestas$^{1,2}$%
  \thanks{$^1$Robotics Institute, Athena Research and Innovation Center, Maroussi 15125, Greece. }%
  \thanks{$^2$School of Electrical and Computer Engineering, National Technical University of Athens, 15773 Athens, Greece. }%
  \thanks{$^{\ast}$Equally contributing authors. }%
  \thanks{Email: \href{mailto:oikonpar@mail.ntua.gr}{\nolinkurl{oikonpar@mail.ntua.gr}}}
}

\begin{document}

\def\cond{False}

\maketitle
\thispagestyle{empty}
\pagestyle{empty}

\begin{abstract}

The perception of an object’s surface is important for robotic applications enabling robust object manipulation.
The level of accuracy in such a representation affects the outcome of the action planning, especially during tasks that require physical contact, e.g. grasping.
In this paper, we propose a novel iterative method for 3D shape reconstruction consisting of two steps. At first, a mesh is fitted on data points acquired from the object's surface, based on a single primitive template. Subsequently, the mesh is properly adjusted to adequately represent local deformities.
Moreover, a novel proactive tactile exploration strategy aims at minimizing the total uncertainty with the least number of contacts, while reducing the risk of contact failure in case
the estimated surface differs significantly from the real one.
The performance of the methodology is evaluated both in 3D simulation and on a real setup.

\end{abstract}

\section{Introduction}\label{sec:intro}

Perception is a fundamental aspect in robotics; it is the ability of the robot to gain knowledge of the objects, the surroundings, the environment dynamics and other agents. Such knowledge is important for setting a goal, planning the proper high-level actions, and executing them.
Shape reconstruction is one of the processes that are closely related to perception, and it aims at creating a representation of an object in the robot's environment. Such comprehension is important for robots, since it allows them to interact with the object in many different ways, e.g. in grasping tasks.

In such applications, data acquisition is crucial. Some methods perform shape reconstruction using only visual data, in the format of RGB images or point-clouds. While this approach enables fast reconstruction due to the large amount of data from a single capture, it lacks in many aspects. For instance, the sensitivity of camera sensors to light conditions might introduce noise, providing only a coarse shape approximation. Additionally, the lack of physical contact makes it unsuitable for deformable objects that require physical interaction to estimate their shape and dynamic properties. With a fixed viewpoint, only a partial view of the object is available due to occlusions.
Despite the accuracy of shape reconstruction in haptic-only applications, the sparsity of surface points slows convergence due to the multiple contacts required. 
Humans naturally combine both approaches in a multi-modal framework~\cite{ernst2002humans}.

The exploration strategy and the processing of sensory data to obtain a 3D object representation are key concerns in robotics.
In this work, we propose a methodology for reconstructing an unknown object's surface by efficiently exploiting surface points obtained through active tactile exploration, using only a minimal visual prior of a small surface portion as initial knowledge. The approach follows an iterative two-step process: first, a mesh is coarsely fitted using a primitive template, then refined to capture local deformities.
Here, the active tactile exploration strategy plays pivotal role attempting to minimize the number of contacts required for providing an accurate shape representation, while operating proactively to avoid contact failures.
Moreover, bounded score values are used to indicate
the level of uncertainty of the estimation across the object's surface.
Such scores are propagated by the actual visual/tactile points whose uncertainties denote the level of reliability of the sensors' measurements.
The methodology is evaluated in simulation for various object shapes, demonstrating fast 3D reconstruction while avoiding contact failures. Preliminary results are also obtained from real-world execution.


The contributions of our work are summarized below:
\begin{itemize}
  \item a single primitive template guides the exploration;
  \item two-step coarse-to-fine fitting: estimation of global template's features followed by local fitting;
  \item uncertainties in sensory measurements: not all obtained surface points are equally reliable;
  \item two-fold exploration strategy: minimize required contact points while diminishing the risk of contact failure.
\end{itemize}


\section{Related Work}\label{sec:relwork}

The relevant bibliography could be classified into three main classes:
visual-only, haptic-only, and visuo-haptic applications.
Aiming at shape completion by visual-only information, the most recent approaches employ data-driven solutions, such as convolutional neural networks \cite{park2019deepsdf,varley2017shape,lundell2019robust}, while others prefer modest optimization techniques, such as primitive-based fitting to point-cloud
\cite{schnabel2009completion}.
On the other hand, the object shape reconstruction using haptic-only means is a common approach in robotics \cite{seminara2019active}.
In a significant portion of these works, e.g. \cite{matsubara2017active,jamali2016active,yi2016active,driess2017active,luo2016iterative}, the interaction between the robot and the object generates a single or sparse surface points on the local surface at each iteration.
This is the main drawback of such applications since it requires multiple contact instances, resulting in slow reconstruction convergence.
To cope with the sparsity of the surface points,
other approaches enhance the process with additional factors; for instance, \cite{abraham2017ergodic} exploits both contact points and noncontact motion to accelerate the reconstruction, while in \cite{sommer2016multi} and \cite{khadivar2023online} a multi-fingered robotic hand gathers surface data with multiple contact receptors.
The last category
incorporates both haptic and visual data \cite{navarro2023visuo}.
Here, e.g. in \cite{bjorkman2013enhancing,wang20183d,gropp2020implicit,tahoun2021visual,suresh2022shapemap,rustler2022active}, the common aspect is that the exploitation of the information takes place in two separate steps: a prediction of the object's shape is initially derived from visual priors, while the estimation is gradually improved through tactile exploration.
In the approaches of the last class, the visual part contributes the most to the object reconstruction through shape completion, while only few tactile points are required for refining the estimation, mainly in the occluded regions. For this to be accomplished, it is necessary for the visual data to concern a significantly large portion of the total surface. But what if the information from the visual source covers only a small part of the object due to e.g. occlusions from other objects or non-ideal viewing angle, hence the shape completion is not feasible? In this work, the latter case is considered.

Another important factor that varies among works is the object shape and uncertainty representation method. From this aspect, these works may be distinguished in two large classes based on the type of representation: discrete and continuous.
At the first one, the most popular approaches exploit
mesh-based 3D shape reconstruction models \cite{varley2017shape,smith2021active,smith20203d}, and voxels as used in computer graphics \cite{lundell2019robust,wang20183d,tahoun2021visual,hongyu2023vihope}.
As regards the second class, a portion of the proposed implementations suggests the employment of continuous signed distance functions (SDFs), as in \cite{suresh2022shapemap}.
In the literature, SDF-based shape representations have been derived in several ways, e.g. from data-driven approaches \cite{park2019deepsdf} or through implicit geometric regularization \cite{gropp2020implicit} using the obtained surface points as boundary condition \cite{rustler2022active}.
In \cite{dragiev2011gaussian}, the authors introduced the use of Gaussian Process Implicit Surfaces (GPIS) \cite{williams2007gaussian} for modelling the surface and the associated shape uncertainty by fusing sensory information from various sources. Since then, GPIS has become a common approach for object shape representation and has been widely used in several works \cite{matsubara2017active,driess2017active}.
A probabilistic representation is also attempted in \cite{jamali2016active,yi2016active,bjorkman2013enhancing}.
In this work, it is assumed that a wide variety of shapes and objects could be seen as mildly or severely deformed ellipsoids.
The formation of the estimated surface in ellipsoid shape may be crucial for the reconstruction at the early stages where only a few surface points have been acquired, providing a coarse approximation of the object's real size and center location, and hence facilitating tactile exploration.
The 3D mesh representation allows such object formation without being restricted by the absence of surface points on its unexplored regions.

\begin{figure}[t]
\begin{center}
\includegraphics[width=0.99\linewidth]{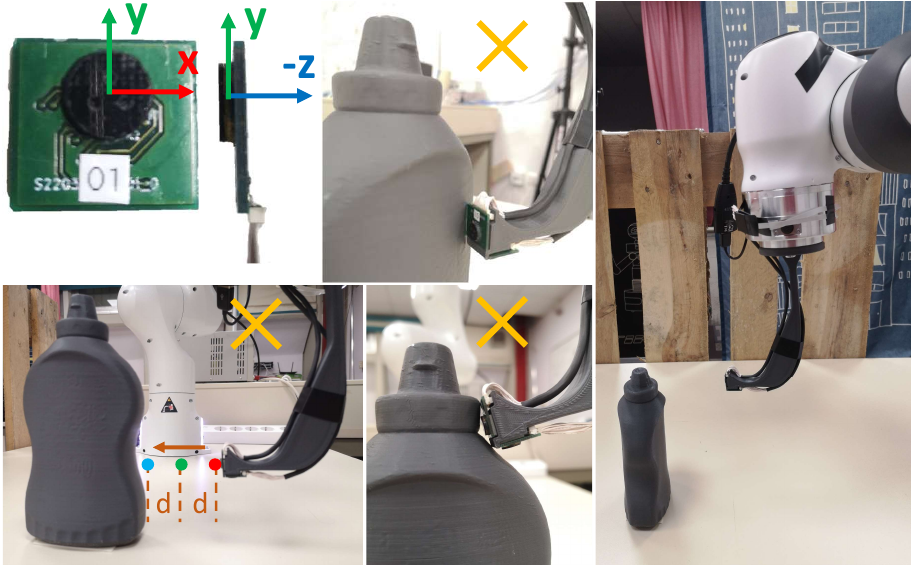}
\end{center}
\vspace{-0.4cm}
\caption{\small{
\textbf{Top left:}
The tactile sensor is a variant of the Shokac Chip 6DoF-P18 by Touchence~\cite{touchsense_shokac} and measures the force/torque applied along all three axes.
\textbf{Bottom left:} At each iteration, the robot moves within a predefined distance $d$ before (red) and after (blue) the candidate contact point (green) along its estimated normal vector. In case of no contact, the attempt is considered as failure. \textbf{Rest with ``X":} Other cases of contact failure if the sensing part (black surface) does not directly touch the object. \textbf{Right:} The real experimental setup.
}}
\vspace{-0.7cm}
\label{fig:tsensor}
\end{figure}

The tactile exploration strategy is another aspect of distinction among the haptic-related works, and is highly considered in this paper. The simplest way to physically explore a surface is through uniform sampling \cite{suresh2022shapemap} where the approach points are fixed a priori.
However, in general, a more sophisticated approach is preferred, where the haptic data are gathered incrementally based on
an active exploration strategy.
The common objective is the 
reduction of the shape uncertainty with the fewest possible 
actions, i.e. the minimization of the
number of contact points
\cite{yi2016active,bjorkman2013enhancing,wang20183d,rustler2022active}.
In \cite{wang20183d}, they select the
touch location based only on the uncertainty of the area that the tactile sensor may cover.
However, considering also the neighboring region, as in the present work, may also accelerate reconstruction.
Additional objectives may be included
in the optimization, such as minimizing the travel cost between consecutive contact points \cite{matsubara2017active}.
Data-driven approaches have also been attempted \cite{smith2021active}. In \cite{fleer2020learning}, the exploration is learned as a sequential decision-making process within a reinforcement learning framework by rewarding the correct object classification.
Dynamic shape exploration frameworks have also been proposed either using query paths \cite{driess2017active} or reinforcement learning \cite{jiang2022active,shahidzadeh2024actexplore}, where the robot gathers data by sliding on the object's surface. However, such approaches have the disadvantage of slow exploration since they take close steps for scanning the whole surface. Alternative strategies \cite{sommer2016multi,driess2019active} use multiple contact points for collecting data more efficiently.
A factor that is not considered in the aforementioned exploration strategies is the risk of contact failure. Such cases are depicted in Fig.~\ref{fig:tsensor}. Considering that the computation of the next surface point is derived based on the estimated - and not the real - shape, it might guide the exploration to a region that is far from the real surface. Such action might induce a wasted motion that decelerates the reconstruction, or an undesired interaction with the object's surroundings. Hence, here the minimization of the risk of contact failure is highly considered.

\section{Methodology}\label{sec:meth}



\begin{figure*}[t]
    \centering
\includegraphics[width=.90\linewidth]{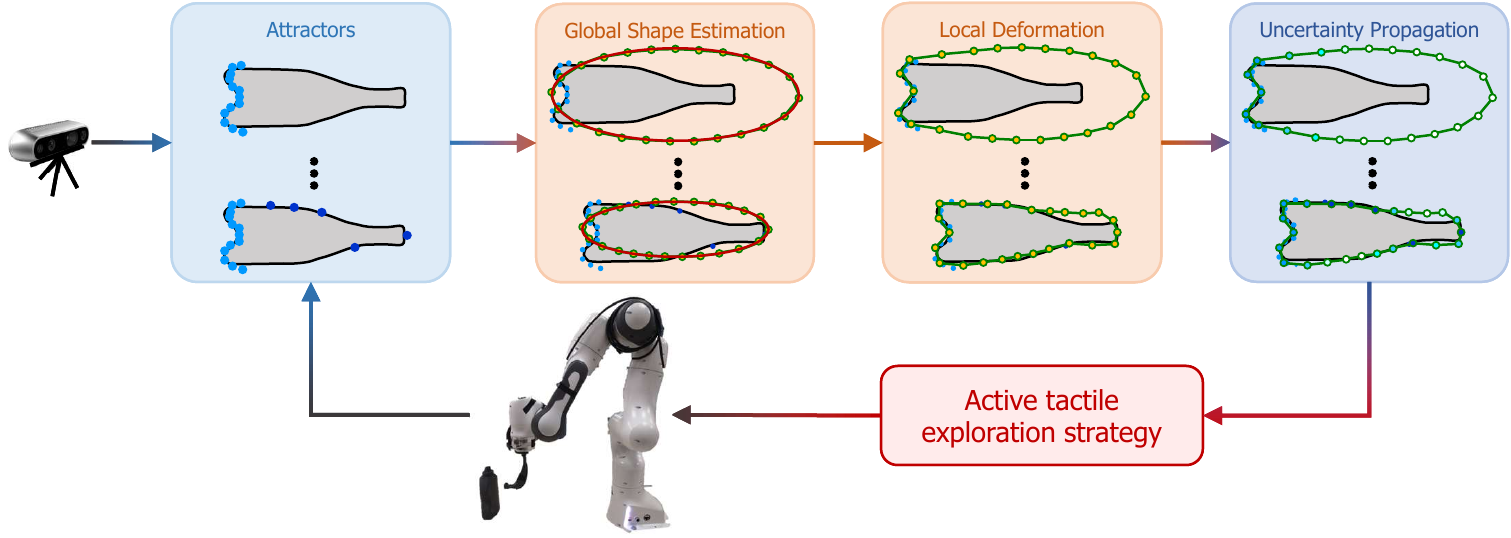}
\vspace{-0.2cm}
\caption{\small{
Overview of the proposed pipeline. To ease the comprehension of the methodology, the shape reconstruction of a 2d bottle-shape object is considered, where a path graph is used, instead of a 3d mesh. The colored blocks correspond to each of the individual algorithmic components taking part in the reconstruction (details in Section~\ref{sec:meth}). In each block, the top instance illustrates the outcome of the process during the first iteration where only the minimal visual prior (light blue) is available, while the bottom illustration concerns the corresponding outcome when five tactile surface points (dark blue) have been collected.
}}
\vspace{-0.6cm}
\label{fig:block_diagram}
\end{figure*}

This section provides a detailed description of the methodology.
At first, a high level overview is given, which is followed by the analysis of the individual components.

\subsection{Overview}



The considered scenario assumes the existence of visual information obtained by a single-view depth camera, which covers only a small portion of the object's surface
due to e.g. occlusions from other objects or unfavorable viewing angle between the camera and the object of interest.
Given these prior points and the template shape, an initial estimation of the surface is made by calculating the affine transformation of the original template.
Here, only an ellipsoid template is considered and thus the
transformation is the actual parameters of the ellipsoid.
Next, an iterative surface estimation process is performed through the sequential addition of tactile points.
The fitting process, where a mesh is generated, is divided into two sequential discrete stages: (a) the global and (b) the local estimation of the surface (Sections~\ref{sec:globalest}~and~\ref{sec:localest}, respectively).
During this step, each vertex of the mesh is assigned with an uncertainty value
denoting the level at which the local area is considered unexplored (Section~\ref{sec:vertex_uncertainty}).
The outcome of the fitting process is an approximation of the object's surface as a deformed template mesh, i.e. a discrete reconstruction.
Each iteration of the pipeline concludes with the computation of
the next contact point that will be approached by the robot, with the global objective to minimize the total shape uncertainty
and the risk of contact failure, with the fewest possible tactile contacts
(Section~\ref{sec:pselect}).
Fig.~\ref{fig:block_diagram} depicts an overview of the pipeline.



\subsection{Attractors: Visual priors and tactile points}\label{sec:attractors}

All surface points collected by visual and tactile sensory devices are referred to as ``attractors", since they are used for spatially attracting the mesh.
However, before the robot begins to actively explore the object of interest, only visual information is available, i.e. a point cloud captured by a depth camera in fixed viewpoint. In this work, these data correspond to only a small part of the object. Despite insufficient for reconstruction, they are important to spatially locate the object in the 3D workspace of the robot, and derive an approximation of the local surface.
The visual attractors
are assigned with an uncertainty value (Section~\ref{sec:attractor_uncertainty}).
During the tactile exploration, the dataset of attractors is iteratively enriched with surface points.
In this work, the detection of the physical contact with the object is made using the tactile sensor depicted in Fig.~\ref{fig:tsensor}.

\subsection{Uncertainty of attractors}\label{sec:attractor_uncertainty}

A key factor
that affects the shape uncertainty and thus the exploration process, is the uncertainty of attractors. In particular, it is considered that the location of each surface point
derived by the sensors, 
is subject to noise. Thereby, the uncertainty of an attractor implies the level of unreliability of the measurement.
The level of noise is mainly induced by the sensing device itself and it is caused by several factors (e.g. light conditions for depth cameras or miscalibration for force/torque sensors).

A fixed uncertainty value is assigned to the visual attractors, which is normally higher compared to that of the tactile data due to the higher sensitivity of the depth camera.
On the other hand, tactile sensors have sensing parts with non-negligible size, e.g. black surface of the sensor in Fig.~\ref{fig:tsensor}. So, despite the attractor is considered at the center of this sensing part, the real location of the contact may slightly differ.
The instantaneous force and torque measurements during the physical contact with the object may indicate the uncertainty, i.e. the reliability of the obtained surface point.
The uncertainty $u_i$ of the attractor point $i$ is:
\begin{equation}
  u_i = u_{\text{max}} \frac{|T_{x,i}|+|T_{y,i}|}{2 |F_{z,i}| + (|T_{x,i}|+|T_{y,i}|)}
  \label{eq:uncertainty}
\end{equation}
where $u_{\text{max}}$ is the maximum allowed uncertainty for tactile attractors, $T_{x,i}$ and $T_{y,i}$ denote the measured torques on $x$ and $y$ axes, respectively, while $F_{z,i}$ is the force on $z$ axis.
Intuitively, the above equation implies zero uncertainty if the real contact is held at the middle of the sensing part where $T_{x,i}=T_{y,i}=0$, and $u_{\text{max}}$ if it takes place at its edge.

\subsection{Template-based global shape estimation}\label{sec:globalest}

Locally changing nodes of a mesh surface can lead to a very slow procedure.
Assuming an underlying structure
can greatly reduce the required steps towards a rough estimation of the pose and the shape of the object.
The ellipsoid assumption is a generic enough template that can guide the pose of arbitrary objects. Implementation-wise, we selected a gradient-descent approach which optimizes a least squares loss of the ellipsoid formula over the set of rotation $\mathbf{R}$, translation $\mathbf{t}$ and scale $\mathbf{s}$ parameters. Given the attractor points $\mathbf{x}_i$, this can be written as:
\begin{align*}
    \mathbf{y_i} =  (\mathbf{x_i} - \mathbf{t}) \, \text{diag}{(\mathbf{s})} \, \mathbf{R}\\
    \underset{\mathbf{R}, \mathbf{t}, \mathbf{s}}{\text{minimize}}  \sum_i (\mathbf{y}_i {\mathbf{y}_i}^\intercal - 1)^2
\end{align*}

An ellipsoid can be analytically estimated through the aforementioned least squared formulation, but such solution may result to non-practical overestimations of the shape in the first steps due to the existence of only a few visual points. 
On the other hand, the gradient-descent approach results to underestimation due a ``modest" initialization. 
Moreover, gradient-based solutions can handle arbitrary templates for potential extension of the framework. 
Eventually, the generated solution fits the template to the existing attractors and ``builds" a coarse mesh representation of the surface. 
The surface details will be acquired from the upcoming step.







\subsection{Estimating local deformation}\label{sec:localest}

The optimization of the object's global shape estimation is followed by the local modulation of the mesh to properly fit on the attractors.
The goal of this process is to compute the displacement of each vertex of the mesh based on its distance from the surrounding attractors.

This step relies on a radial basis function interpolation over the vertex points. Specifically, a thin plate spline variant is used in order to compute a displacement value $d_i$ for a set of points $\mathbf{p}_i$, belonging to the global mesh, with normal vectors $\mathbf{n}_i$, such that the actual attractor points $\mathbf{x}_i$ are represented as:
\begin{equation}
  \mathbf{x}_i  = \mathbf{p}_i + d_i \mathbf{n}_i
  \label{eq:displacement}
\end{equation}
In other words, the interpolation function $\mathit{F}: \mathbb{R}^3 \rightarrow  \mathbb{R}$  will be fitted to satisfy $\mathit{F}(\mathbf{p}_i) = d_i$. Displacements $d_i$ are positive for bumps or negative for dents in the global surface.

To estimate this interpolation function, we must first detect the mesh points $\mathbf{p}_i$ that satisfy Eq.~\ref{eq:displacement}. 
These points are not necessarily vertices of the mesh, but the projections of the attractors on the global mesh,
i.e. mesh points of the shortest distance from the attractors.
Implementation-wise, for each attractor $\mathbf{x}_i$, we seek for the closest mesh triangle that includes the attractor's projection. The corresponding normal $\mathbf{n}_i$ is the selected triangle's normal.
After defining the interpolator $\mathit{F}$, we can estimate the actual mesh displacements on the vertices $\{  \mathbf{v}_i  \}$, where the normals are known.
The new, locally deformed, mesh will be defined by the updated vertices $\mathbf{v}_i + \mathit{F}(\mathbf{v}_i) \mathbf{n}_i$.




\subsection{Uncertainty propagation to vertices}\label{sec:vertex_uncertainty}

A critical information for the reconstructed mesh is the measurement of uncertainty for each vertex. Such uncertainty is useful for the upcoming tactile exploration strategy and can be propagated from the uncertainty of the attractors (Section~\ref{sec:attractor_uncertainty}).
Given an attractor $i$, we find the closest mesh vertex and its connected neighbors according to a traverse threshold (e.g. go as far as 5 edges).
For this set of vertices, we calculate their distance $t$ from the attractor and update their uncertainty by multiplying the uncertainty $u_i$ of the attractor with a proximity weight. Such a weight is defined by a heavy-tailed function $1/(1+t^2)$ in order to propagate low uncertainty values to farther neighborhoods. For vertices that are affected from multiple attractors, the minimum uncertainty is kept.


\begin{figure}
    \centering
    \includegraphics[width=.8\linewidth]{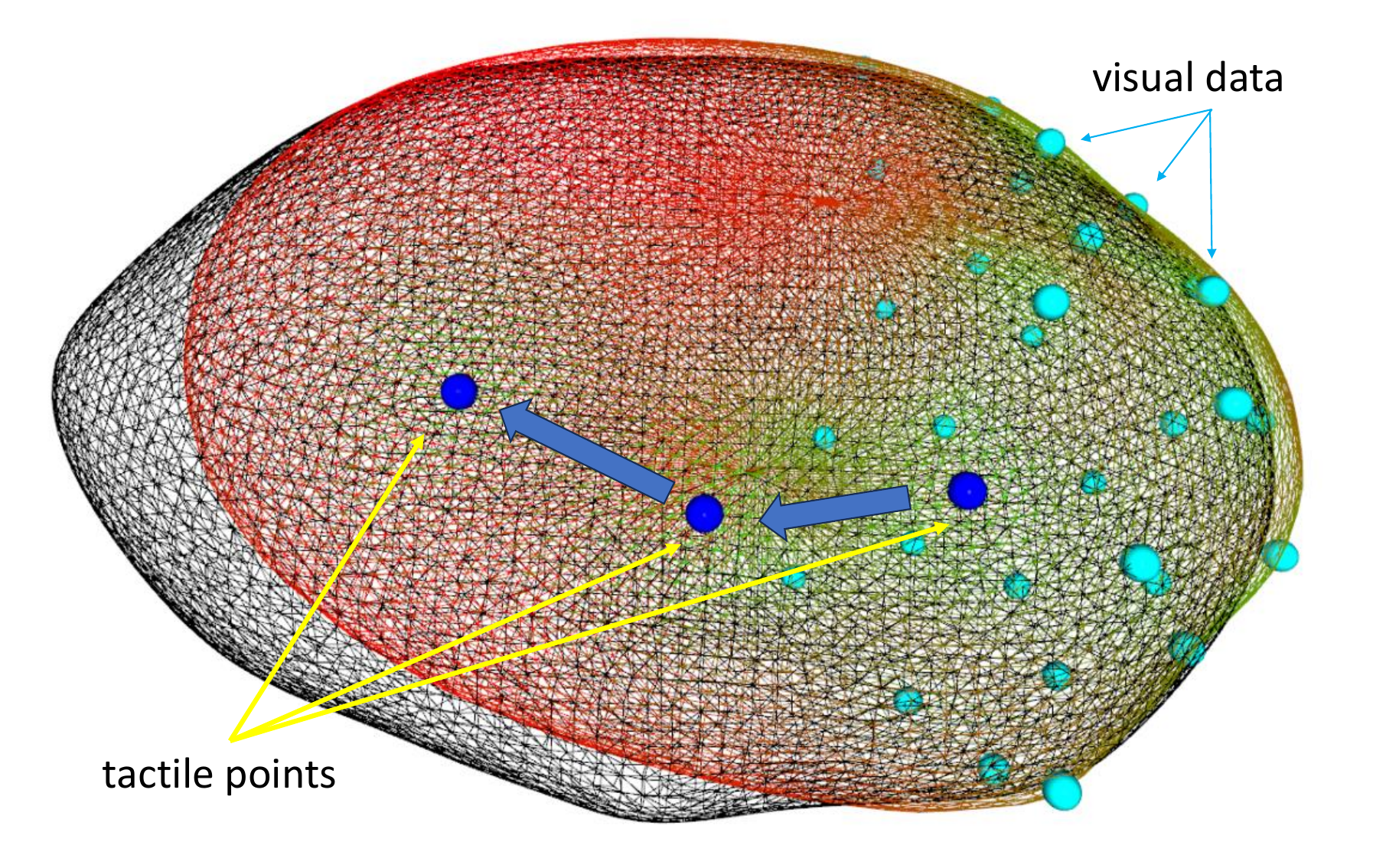}
    \vspace{-.1cm}
    \caption{\small{
    First three iterations of the proposed active tactile exploration strategy (Section~\ref{sec:pselect}). At the side of the visual points (cyan), the reconstructed mesh (red-green) fits the actual one (grey). At the other side tactile exploration is required. Following the proposed algorithm, a path of tactile points (blue) is created from a neighborhood close to the visual data up to the unexplored side of the surface, resembling a depth first search.
    }}
    \vspace{-.6cm}
    \label{fig:point_selection}
\end{figure}


\subsection{Active tactile exploration strategy}\label{sec:pselect}



The improvement of the surface estimation requires continuous and active tactile exploration.
A strategy that optimally selects the contact points is required, so that
the estimation error is decreased with the highest possible rate with respect to the number of contacts.
Under the scope of optimization, if
minimizing the total uncertainty is the only goal, then a typical solution
adopted by related works (Section~\ref{sec:relwork}),
is the selection of the vertex with the maximum uncertainty at each iteration.
However, the candidate approach points are computed based on the estimated surface, and hence they do not necessarily lie on the real one.

For instance, at the first iteration, where only visual attractors are available
, the surface estimation is not reliable, e.g. an overestimation or underestimation of the object's size might occur, (Fig.~\ref{fig:reconstruction_progress} - last row, second column).
In such a case, the vertex of highest uncertainty might be located in an area where no real surface exist close to, resulting in contact failure when approached by the robot.


To mitigate the risk of contact failure while ensuring fast shape estimation, a two-fold active tactile exploration strategy is proposed.
Intuitively,
the exploration attempts to take ``cautious" steps, focusing on regions that are close to vertices of low uncertainty.
However, only this constraint would be akin to a breadth first search, which would delay the exploration.
In contrast, a depth first exploration is preferred in order to rapidly discover the global structure of the object and eventually reduce the total uncertainty, as seen in Fig.~\ref{fig:point_selection}.
Implementation-wise, the selected point $j$ results from an optimization problem which seeks for the vertex with the highest mean geodesic distance $G_j$ from the ``confident" vertices of its neighborhood, and the highest uncertainty score $U_j$. These two factors are properly weighted in order to promote either the conservative or the aggressive exploration.
Moreover, two additional constraints are set: the uncertainty of the candidate vertex exceeds a predetermined threshold $u'_{\text{min}}$, while at least one of its adjacent vertices is ``confident", i.e. with uncertainty below $u'_{\text{max}}$.
Mathematically, for the candidate contact point $w$ the following equations must hold:
\begin{equation}
\label{eq:geodesic}
\begin{aligned}
    w = \text{argmax}_j \left( \alpha_G G_j + \alpha_U U_j \right) \\
    \text{min} \{ u_i \}_w \leq u'_{\text{max}} \\
    u_w \geq u'_{\text{min}}
\end{aligned}
\end{equation}
where $j=\{1,2,\dots\}$ are all vertices, $\{\alpha_G,\alpha_U\} \in \mathbb{R}^+$ are the weighting parameters, $u_w$ is the uncertainty of vertex $w$, and $\{ u_i \}_w$ is the set of uncertainties of all $w$'s neighbors.

\section{Experimental Evaluation}\label{sec:exp}


\subsection{3D reconstruction of various objects}

\begin{figure*}[t]
    \centering
    \includegraphics[width=.90\linewidth]{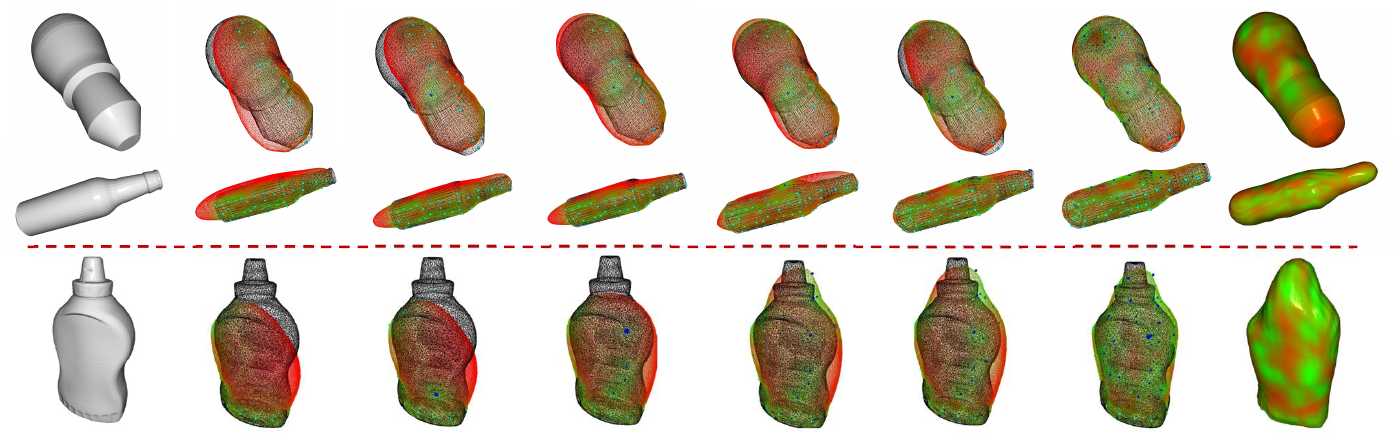}
    \vspace{-.3cm}
    \caption{\small{
    Iterative progression of the proposed methodology. The first two rows corresponds to the evaluation performed in simulation, while the last one took place on the real experimental setup (Sec.~\ref{sec:real_exp}).
    From left to right: target mesh, 6 instances of the reconstruction at 0 (initial), 1, 2, 5, 10 and 30 iterations and the resulting mesh after 30 iterations. 
    The intermediate instances depict the predicted surface overlapped on the target shape, where green and red regions denote low and high uncertainty, respectively.
    Visual points are depicted with cyan color, while tactile points are visualized as blue points (zoom is required).
    }}
    \vspace{-.3cm}
    \label{fig:reconstruction_progress}
\end{figure*}

Initial experimentation was performed in simulation where different mesh objects are considered for reconstruction:
a lamp, a wine bottle and several objects drawn from the YCB dataset~\cite{calli2015ycb}; namely the pear, the mini soccer ball, the coffee can, the mustard container and the box of sugar.
Fig.~\ref{fig:reconstruction_progress} contains instances of the progress of the proposed iterative algorithm at each column, as well as the final estimated surface, while each row corresponds to an object; the first two objects are examined here.
It is seen that for the lamp - that is similar to the template - the surface rapidly converge close to the targeted one, while the wine bottle has been proven more challenging.
That is to be expected since a more complex structure should be captured through a few points, when no other prior is utilized (e.g. symmetry constraints).

\begin{figure}[t]
    \centering
    \includegraphics[width=.99\linewidth]{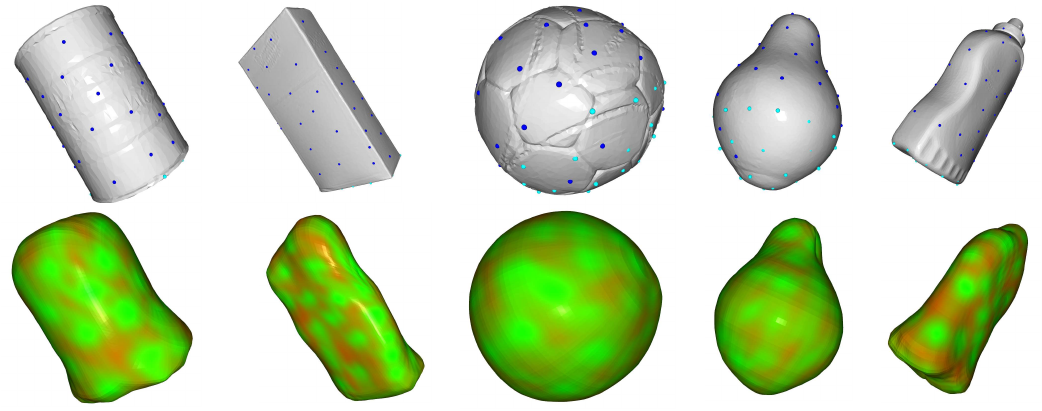}
    \vspace{-.5cm}
    \caption{\small{
    Reconstruction of several objects (from left to right: coffee can, box of sugar, mini soccer ball, pear, mustard container) from YCB dataset, after 50 iterations of the methodology (including contact failure iterations). \textbf{Top:} Real mesh of the object and collected attractors (visual in cyan, tactile in blue). \textbf{Bottom:} Reconstructed shape colored with uncertainty (high in red, low in green).
    }}
    \vspace{-0.75cm}
    \label{fig:ycb_reconstruction}
\end{figure}

The outcome of the proposed shape reconstruction after 50 iterations for some objects of YCB dataset is illustrated in Fig.~\ref{fig:ycb_reconstruction}. The estimated meshes have managed to capture the global shape of the objects with only a small number of surface points. As expected, the fewer number of edges, the better the reconstruction, e.g. ball and pear, while even in some extreme cases, e.g. box of sugar, the reconstructed mesh constitutes a qualitative representation of the object.
The distribution of the colored uncertainty implies that even for a small number of tactile points the proposed exploration strategy have achieved to uniformly explore all regions of the objects' surfaces, without imposing such a requirement in the designed strategy.
Indicating areas with high uncertainty (red) does not necessarily imply that the reconstructed local surface varies significantly from the real one, but that there are no visual/tactile attractors obtained in the specific region. There, the local surface estimation has been derived based on the reconstruction of adjacent regions rather than from actual information (attractors).
A quantification of the reconstruction's accuracy accompanied by the number of contact failures, for all considered objects is given in Table~\ref{tab:comparison}.

\begin{figure}[t]
    \centering
    \includegraphics[width=.99\linewidth]{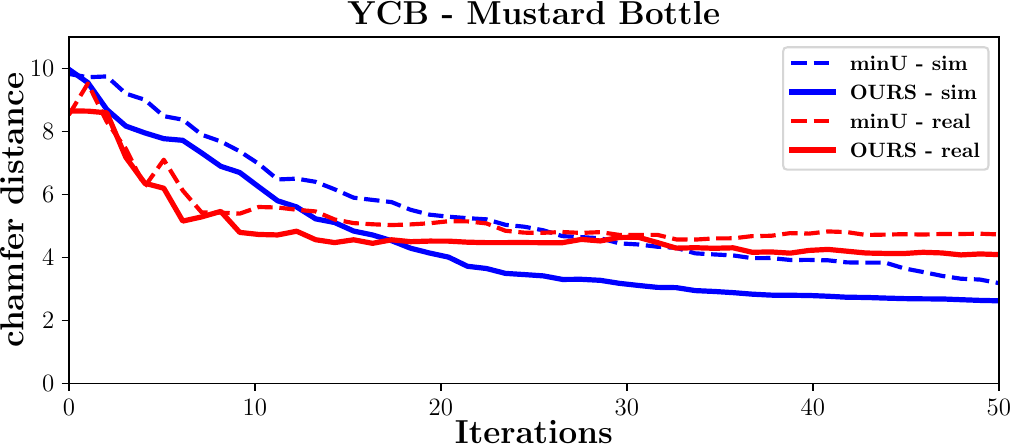}
    \vspace{-.5cm}
    \caption{\small{
    Comparison of the two exploration strategies (OURS vs. minU) as regards the evolution of chamfer distance over time in both simulation and real experimental setup, for a specific object. Note that, the depicted plots correspond to the mean of chamfer distances after five independent executions of each case.
    }}
    \vspace{-.7cm}
    \label{fig:mustard_bottle_comparison}
\end{figure}

\begin{table*}[t!]
  \centering
  \caption{\small{
  Quantitative comparison between the proposed exploration strategy (OURS) and the common approach that targets at only minimizing the total uncertainty (minU). Two metrics have considered: (a) the chamfer distance (in mm) between the real and the reconstructed point cloud and (b) the number of contact failures after 50 iterations. The lower the metrics, the more efficient each strategy is. The first two columns indicate the features and the strategies
  considered in the comparative study,
  while the rest ones provide the mean and standard deviation
  after three executions.
  The last column concerns the experiment on the real robot.
  }}
    \begin{tabular}{c|c||c|c|c|c|c|c|c||c}
    \toprule
    \textbf{Metric} & \textbf{Strategy} & \textbf{lamp} & \textbf{wine b.} & \textbf{coffee can} & \textbf{sugar box} & \textbf{ball} & \textbf{pear} & \textbf{mustard c.} & \textbf{mustard c. (R)} \\
     
    \midrule
    & \textbf{OURS} & 1.07 ± 0.01 & 3.05 ± 0.19 & 3.89 ± 0.15 & 3.82 ± 0.12 & 1.92 ± 0.01 & 1.17 ± 0.0 & 2.62 ± 0.13 & 4.1 ± 0.13 \\
    \cmidrule{2-10}
    \textbf{chamfer} & \textbf{minU} & 1.08 ± 0.01 & 3.47 ± 0.33 & 7.56 ± 1.62 & 9.97 ± 3.74 & 1.93 ± 0.02 & 1.78 ± 0.82 & 3.18 ± 0.93 & 4.73 ± 0.36 \\
    \midrule
    & \textbf{OURS} & 1.0 ± 0.7 & 12.3 ± 1.8 & 13.3 ± 0.9 & 12.7 ± 1.8 & 3.7 ± 1.1 & 0.0 ± 0.0 & 4.1 ± 1.9 & 9.1 ± 2.1 \\
    \cmidrule{2-10}
    \textbf{\#failures} & \textbf{minU} & 6.7 ± 3.1 & 26.25 ± 1.9 & 29.7 ± 6.2 & 32.7 ± 7.8 & 19.0 ± 4.7 & 14.7 ± 16.2 & 11.3 ± 9.2 & 18.3 ± 5.5 \\
    \bottomrule
    \end{tabular}%
  \label{tab:comparison}%
\vspace{-0.5cm}
\end{table*}



\subsection{Impact of the proposed tactile exploration strategy} \label{sec:comparison}

A key aspect of the methodology is the proactive tactile exploration strategy which focuses on minimizing both the global uncertainty and the risk of contact failure.
Its performance is evaluated, and compared in terms of number of contact failures and reconstruction error, with the common approach - namely, minU - that only aims at minimizing the global uncertainty.
To ensure a fair comparison, the proposed reconstruction methodology is executed twice for the same object, and under the same parameters and initial
conditions, with the only difference relying in the tactile exploration strategy.
The experiments took place both in simulation and on the real setup.
In this set of experiments, contact failure is considered if
the following state is met:
the euclidean distance between the locations of the candidate point on the estimated surface and the corresponding point on the real one exceeds a threshold - here, $15$ mm.
The latter point
is derived from the intersection with the real surface, of the line defined by the candidate point and its normal on the estimated surface. 
For both strategies, in case of contact failure the shape estimation is not updated hence the reconstruction error remains the same for one iteration.

The outcomes of the execution for all objects are provided in Table~\ref{tab:comparison}. It is evident that for the same number of contact attempts, the proposed strategy manages to achieve a considerably lower reconstruction error with respect to minU for the most challenging cases, i.e. coffee can, sugar box and mustard container, due to the significantly reduced number of contact failures. Moreover, the deviation of the latter among the several executions remains low, highlighting the consistency of our approach.
A flaw of the minU strategy is that, at some executions for certain objects (e.g. sugar box) it was trapped into an infinite loop of consecutive contact failures. However, the results presented in Table~\ref{tab:comparison} concern only the cases where minU strategy managed to make relatively significant progress in reconstruction. Along the same lines, Fig.~\ref{fig:mustard_bottle_comparison} enhances the aforementioned outcomes.

\subsection{Experiments on real setup}
\label{sec:real_exp}

Here, an overview of the experimental procedure and its outcomes on the real robot are given.
The visual priors were captured using a single-view RGB-D camera (Intel RealSense d435i).
For the tactile exploration, a custom 3D printed mount was attached at the tip of a 7-DoF robotic arm (Franka Emika Panda), carrying two tactile sensors (Fig.~\ref{fig:tsensor} in different orientations.
As regards the object under consideration, a 3D printed replica of the mustard container included in the YCB dataset is used.
Note that, the measurements obtained from all components, e.g. point-cloud data, tactile sensor's location, are expressed to the Panda robot's base frame after applying the proper transformations.

Initially,
the camera captures a small part of the object's surface 
and a fixed uncertainty score of $0.4$ is assigned to all visual attractors, implying the noisy source of their measurement.
Subsequently, the proposed methodology is applied (Section~\ref{sec:meth}).
At the end of each iteration, the tactile sensor that will be used for detecting the physical contact with the object is selected based on the estimated normal vector at the candidate contact point from the reconstructed surface.
Then, the robot slowly moves the sensor along this vector, within a certain distance around the candidate contact point (see Fig.~\ref{fig:tsensor}). In case of non-zero force/torque measurements, the contact location is added to the set of attractors with uncertainty computed by Eq.~\ref{eq:uncertainty}. Otherwise, the iteration is considered as contact failure.

The evolution of the shape reconstruction is illustrated in the last row of Fig.~\ref{fig:reconstruction_progress}.
The final reconstruction is good enough, despite the noise induced by the visual sensor which has slowed down the procedure by complicating the global estimation step and has created a surface bump (left side of the mesh).
Correcting such mistakes requires some tactile points to be taken near the visual attractors. Since visual points are assigned a higher uncertainty compared to tactile ones, such actions are possible but require multiple steps (i.e., to fully reduce uncertainty on every other region).








\section{Future Work}\label{sec:futwork}

The proposed methodology offers some space for improvement in various aspects. For instance, the reconstruction could be accelerated by incorporating repulsors, apart from attractors, to indicate free-space at the locations where the robot has travelled and no contact has been perceived, with the aim to restrict the estimated surface. Moreover, some model-based modules (e.g. exploration strategy) could be replaced by learning-based ones, with the aim to avoid the hyperparameter tuning, and ensure robustness.
Additional optimization criteria, such as travel cost, could speed up real-world reconstruction. Eventually, a template selection strategy could choose among multiple templates in real-time.

\bibliographystyle{./bibliography/IEEEtran}
\bibliography{./bibliography/IEEEabrv,./bibliography/ICRA_2024}

\begin{thebibliography}{10}
\providecommand{\url}[1]{#1}
\csname url@samestyle\endcsname
\providecommand{\newblock}{\relax}
\providecommand{\bibinfo}[2]{#2}
\providecommand{\BIBentrySTDinterwordspacing}{\spaceskip=0pt\relax}
\providecommand{\BIBentryALTinterwordstretchfactor}{4}
\providecommand{\BIBentryALTinterwordspacing}{\spaceskip=\fontdimen2\font plus
\BIBentryALTinterwordstretchfactor\fontdimen3\font minus \fontdimen4\font\relax}
\providecommand{\BIBforeignlanguage}[2]{{%
\expandafter\ifx\csname l@#1\endcsname\relax
\typeout{** WARNING: IEEEtran.bst: No hyphenation pattern has been}%
\typeout{** loaded for the language `#1'. Using the pattern for}%
\typeout{** the default language instead.}%
\else
\language=\csname l@#1\endcsname
\fi
#2}}
\providecommand{\BIBdecl}{\relax}
\BIBdecl

\bibitem{ernst2002humans}
M.~O. Ernst and M.~S. Banks, ``Humans integrate visual and haptic information in a statistically optimal fashion,'' \emph{Nature}, vol. 415, no. 6870, pp. 429--433, 2002.

\bibitem{park2019deepsdf}
J.~J. Park, P.~Florence, J.~Straub, R.~Newcombe, and S.~Lovegrove, ``Deepsdf: Learning continuous signed distance functions for shape representation,'' in \emph{Proceedings of the IEEE/CVF Conference on Computer Vision and Pattern Recognition (CVPR)}, June 2019.

\bibitem{varley2017shape}
J.~Varley, C.~DeChant, A.~Richardson, J.~Ruales, and P.~Allen, ``Shape completion enabled robotic grasping,'' in \emph{2017 IEEE/RSJ international conference on intelligent robots and systems (IROS)}.\hskip 1em plus 0.5em minus 0.4em\relax IEEE, 2017, pp. 2442--2447.

\bibitem{lundell2019robust}
J.~Lundell, F.~Verdoja, and V.~Kyrki, ``Robust grasp planning over uncertain shape completions,'' in \emph{2019 IEEE/RSJ International Conference on Intelligent Robots and Systems (IROS)}.\hskip 1em plus 0.5em minus 0.4em\relax IEEE, 2019, pp. 1526--1532.

\bibitem{schnabel2009completion}
\BIBentryALTinterwordspacing
R.~Schnabel, P.~Degener, and R.~Klein, ``Completion and reconstruction with primitive shapes,'' \emph{Computer Graphics Forum}, vol.~28, no.~2, pp. 503--512, 2009. [Online]. Available: \url{https://onlinelibrary.wiley.com/doi/abs/10.1111/j.1467-8659.2009.01389.x}
\BIBentrySTDinterwordspacing

\bibitem{seminara2019active}
L.~Seminara, P.~Gastaldo, S.~J. Watt, K.~F. Valyear, F.~Zuher, and F.~Mastrogiovanni, ``Active haptic perception in robots: a review,'' \emph{Frontiers in neurorobotics}, vol.~13, p.~53, 2019.

\bibitem{matsubara2017active}
T.~Matsubara and K.~Shibata, ``Active tactile exploration with uncertainty and travel cost for fast shape estimation of unknown objects,'' \emph{Robotics and Autonomous Systems}, vol.~91, pp. 314--326, 2017.

\bibitem{jamali2016active}
N.~Jamali, C.~Ciliberto, L.~Rosasco, and L.~Natale, ``Active perception: Building objects' models using tactile exploration,'' in \emph{2016 IEEE-RAS 16th International Conference on Humanoid Robots (Humanoids)}, 2016, pp. 179--185.

\bibitem{yi2016active}
Z.~Yi, R.~Calandra, F.~Veiga, H.~van Hoof, T.~Hermans, Y.~Zhang, and J.~Peters, ``Active tactile object exploration with gaussian processes,'' in \emph{2016 IEEE/RSJ International Conference on Intelligent Robots and Systems (IROS)}, 2016, pp. 4925--4930.

\bibitem{driess2017active}
D.~Driess, P.~Englert, and M.~Toussaint, ``Active learning with query paths for tactile object shape exploration,'' in \emph{2017 IEEE/RSJ International Conference on Intelligent Robots and Systems (IROS)}, 2017, pp. 65--72.

\bibitem{luo2016iterative}
S.~Luo, W.~Mou, K.~Althoefer, and H.~Liu, ``Iterative closest labeled point for tactile object shape recognition,'' in \emph{2016 IEEE/RSJ International Conference on Intelligent Robots and Systems (IROS)}, 2016, pp. 3137--3142.

\bibitem{abraham2017ergodic}
I.~Abraham, A.~Prabhakar, M.~J. Hartmann, and T.~D. Murphey, ``Ergodic exploration using binary sensing for nonparametric shape estimation,'' \emph{IEEE robotics and automation letters}, vol.~2, no.~2, pp. 827--834, 2017.

\bibitem{sommer2016multi}
N.~Sommer and A.~Billard, ``Multi-contact haptic exploration and grasping with tactile sensors,'' \emph{Robotics and autonomous systems}, vol.~85, pp. 48--61, 2016.

\bibitem{khadivar2023online}
F.~Khadivar, K.~Yao, X.~Gao, and A.~Billard, ``Online active and dynamic object shape exploration with a multi-fingered robotic hand,'' \emph{Robotics and Autonomous Systems}, vol. 166, p. 104461, 2023.

\bibitem{navarro2023visuo}
N.~Navarro-Guerrero, S.~Toprak, J.~Josifovski, and L.~Jamone, ``Visuo-haptic object perception for robots: an overview,'' \emph{Autonomous Robots}, vol.~47, no.~4, pp. 377--403, 2023.

\bibitem{bjorkman2013enhancing}
M.~Bj{\"o}rkman, Y.~Bekiroglu, V.~H{\"o}gman, and D.~Kragic, ``Enhancing visual perception of shape through tactile glances,'' in \emph{2013 IEEE/RSJ International Conference on Intelligent Robots and Systems}.\hskip 1em plus 0.5em minus 0.4em\relax IEEE, 2013, pp. 3180--3186.

\bibitem{wang20183d}
S.~Wang, J.~Wu, X.~Sun, W.~Yuan, W.~T. Freeman, J.~B. Tenenbaum, and E.~H. Adelson, ``3d shape perception from monocular vision, touch, and shape priors,'' in \emph{2018 IEEE/RSJ International Conference on Intelligent Robots and Systems (IROS)}.\hskip 1em plus 0.5em minus 0.4em\relax IEEE, 2018, pp. 1606--1613.

\bibitem{gropp2020implicit}
\BIBentryALTinterwordspacing
A.~Gropp, L.~Yariv, N.~Haim, M.~Atzmon, and Y.~Lipman, ``Implicit geometric regularization for learning shapes,'' in \emph{Proceedings of the 37th International Conference on Machine Learning}, ser. Proceedings of Machine Learning Research, H.~D. III and A.~Singh, Eds., vol. 119.\hskip 1em plus 0.5em minus 0.4em\relax PMLR, 13--18 Jul 2020, pp. 3789--3799. [Online]. Available: \url{https://proceedings.mlr.press/v119/gropp20a.html}
\BIBentrySTDinterwordspacing

\bibitem{tahoun2021visual}
M.~Tahoun, O.~Tahri, J.~A. Corrales~Ramón, and Y.~Mezouar, ``Visual-tactile fusion for 3d objects reconstruction from a single depth view and a single gripper touch for robotics tasks,'' in \emph{2021 IEEE/RSJ International Conference on Intelligent Robots and Systems (IROS)}, 2021, pp. 6786--6793.

\bibitem{suresh2022shapemap}
S.~Suresh, Z.~Si, J.~G. Mangelson, W.~Yuan, and M.~Kaess, ``Shapemap 3-d: Efficient shape mapping through dense touch and vision,'' in \emph{2022 International Conference on Robotics and Automation (ICRA)}, 2022, pp. 7073--7080.

\bibitem{rustler2022active}
L.~Rustler, J.~Lundell, J.~K. Behrens, V.~Kyrki, and M.~Hoffmann, ``Active visuo-haptic object shape completion,'' \emph{IEEE Robotics and Automation Letters}, vol.~7, no.~2, pp. 5254--5261, 2022.

\bibitem{smith2021active}
\BIBentryALTinterwordspacing
E.~Smith, D.~Meger, L.~Pineda, R.~Calandra, J.~Malik, A.~Romero~Soriano, and M.~Drozdzal, ``Active 3d shape reconstruction from vision and touch,'' in \emph{Advances in Neural Information Processing Systems}, M.~Ranzato, A.~Beygelzimer, Y.~Dauphin, P.~Liang, and J.~W. Vaughan, Eds., vol.~34.\hskip 1em plus 0.5em minus 0.4em\relax Curran Associates, Inc., 2021, pp. 16\,064--16\,078. [Online]. Available: \url{https://proceedings.neurips.cc/paper_files/paper/2021/file/8635b5fd6bc675033fb72e8a3ccc10a0-Paper.pdf}
\BIBentrySTDinterwordspacing

\bibitem{smith20203d}
E.~Smith, R.~Calandra, A.~Romero, G.~Gkioxari, D.~Meger, J.~Malik, and M.~Drozdzal, ``3d shape reconstruction from vision and touch,'' \emph{Advances in Neural Information Processing Systems}, vol.~33, pp. 14\,193--14\,206, 2020.

\bibitem{hongyu2023vihope}
H.~Li, S.~Dikhale, S.~Iba, and N.~Jamali, ``Vihope: Visuotactile in-hand object 6d pose estimation with shape completion,'' \emph{IEEE Robotics and Automation Letters}, vol.~8, no.~11, pp. 6963--6970, 2023.

\bibitem{dragiev2011gaussian}
S.~Dragiev, M.~Toussaint, and M.~Gienger, ``Gaussian process implicit surfaces for shape estimation and grasping,'' in \emph{2011 IEEE International Conference on Robotics and Automation}, 2011, pp. 2845--2850.

\bibitem{williams2007gaussian}
\BIBentryALTinterwordspacing
O.~Williams and A.~Fitzgibbon, ``Gaussian process implicit surfaces,'' in \emph{Gaussian Processes in Practice}, April 2007. [Online]. Available: \url{https://www.microsoft.com/en-us/research/publication/gaussian-process-implicit-surfaces-2/}
\BIBentrySTDinterwordspacing

\bibitem{fleer2020learning}
\BIBentryALTinterwordspacing
S.~Fleer, A.~Moringen, R.~L. Klatzky, and H.~Ritter, ``Learning efficient haptic shape exploration with a rigid tactile sensor array,'' \emph{PLOS ONE}, vol.~15, no.~1, pp. 1--22, 01 2020. [Online]. Available: \url{https://doi.org/10.1371/journal.pone.0226880}
\BIBentrySTDinterwordspacing

\bibitem{jiang2022active}
S.~Jiang and L.~L. Wong, ``Active tactile exploration using shape-dependent reinforcement learning,'' in \emph{2022 IEEE/RSJ International Conference on Intelligent Robots and Systems (IROS)}, 2022, pp. 8995--9002.

\bibitem{shahidzadeh2024actexplore}
A.-H. Shahidzadeh, S.~J. Yoo, P.~Mantripragada, C.~D. Singh, C.~Fermüller, and Y.~Aloimonos, ``Actexplore: Active tactile exploration on unknown objects,'' in \emph{2024 IEEE International Conference on Robotics and Automation (ICRA)}, 2024, pp. 3411--3418.

\bibitem{driess2019active}
D.~Driess, D.~Hennes, and M.~Toussaint, ``Active multi-contact continuous tactile exploration with gaussian process differential entropy,'' in \emph{2019 International Conference on Robotics and Automation (ICRA)}, 2019, pp. 7844--7850.

\bibitem{touchsense_shokac}
\BIBentryALTinterwordspacing
T.~Inc. Shokac chip 6dof-p18. Accessed on February 25, 2024. [Online]. Available: \url{http://www.touchence.jp/en/company.html}
\BIBentrySTDinterwordspacing

\bibitem{calli2015ycb}
B.~Calli, A.~Singh, A.~Walsman, S.~Srinivasa, P.~Abbeel, and A.~M. Dollar, ``The ycb object and model set: Towards common benchmarks for manipulation research,'' in \emph{2015 International Conference on Advanced Robotics (ICAR)}, 2015, pp. 510--517.

\end{thebibliography}

\end{document}